\documentclass[english]{elsarticle}

\usepackage{subfigure}
\usepackage[latin1]{inputenc}

\title{SlideRunner - A Tool for Massive Cell Annotations in Whole Slide Images}
\author[1]{Marc~Aubreville}
\author[2]{Christof~Bertram}
\author[2]{Robert~Klopfleisch}
\author[1]{Andreas~Maier}

\journal{Bildverarbeitung in der Medizin 2018}

\address[1]{Pattern Recognition Lab, Computer Sciences, Friedrich-Alexander-Universität Erlangen-Nürnberg}
\address[2]{Institute of Veterinary Pathology, Freie Universität Berlin, Germany }

\begin{document}

\maketitle

\begin{abstract}
Large-scale image data such as digital whole-slide histology images pose a challenging task at annotation software solutions. Today, a number of good solutions with varying scopes exist. For cell annotation, however, we find that many do not match the prerequisites for fast annotations. Especially in the field of mitosis detection, it is assumed that detection accuracy could significantly benefit from larger annotation databases that are currently however very troublesome to produce. Further, multiple independent (blind) expert labels are a big asset for such databases, yet there is currently no tool for this kind of annotation available.

 To ease this tedious process of expert annotation and grading, we introduce  SlideRunner, an open source annotation and visualization tool for digital histopathology, developed in close cooperation with two pathologists. 
SlideRunner is capable of setting annotations like object centers (for e.g. cells) as well as object boundaries (e.g. for tumor outlines). It provides single-click annotations as well as a blind mode for multi-annotations, where the expert is directly shown the microscopy image containing the cells that he has not yet rated.

\end{abstract}

\section{Introduction}
Whole-slide images enable automated image analysis in histopathology and is a well-recognized and legally approved method for quantification of immunohistochemistry in research and diagnosis. It has, beyond that, a great potential to further support the pathologists' work in monotoneous, time-consuming tasks such as counting rare events \cite{Bertram:2017kn}.
The availability of large amounts of data has also paved the way for pattern recognition methods that can be used to automatically pre-annotate and analyze whole-slide images. Especially since the advent of deep learning, the need for high-quality data annotation at a large scale has increased significantly, but so have the prospects of high-quality detection results. 

The demand for those large-scale data sets poses two major problems to the annotation task, that are currently unsolved for many medical image recognition domains: The lack in quantity in data, and the lack in quality of labels. 
To overcome the limitations in quantity of expert label data, Albarqouni \textit{et al.} used an aggregation of expert labels and crowdsourcing with non-experts \cite{Albarqouni:2016hy}, showing effectively the power of an enhanced data set. However, it can be stated that expert-labelled data is generally of higher quality than non-expert labelled data.

\subsection*{High rating variance in expert annotations}
Medical images in general, and histology images in particular, often have a high inter-observer variance in rating.  
As Boiesen \textit{et al.} report, subjective differences in grading in breast cancer histology tumor diagnostics can lead to a significantly high variance in the tumor classification \cite{PoulBoiesenParOlaBendahlLola:2009kv}, which in term has implications on an individual, targeted curative treatment. This, of course, calls for computer-based assistance systems, which could significantly contribute to a highly standardized diagnosis and thus could lead to a more uniform classification. 

Further, histo-pathology images are often annotated in an unsuitable way for supervised machine learning on a segmentation task (e.g. just by an arrows pointing at structures, using circles or as per-slide annotation). While this might be a suitable and sufficient annotation for a human observer, it is often too ambiguous for pattern recognition methods. 

\subsection*{Prospects of multi label data sets}

For algorithmic approaches, the quality of input data is a major bottleneck for the quality of the overall outcome. In a machine learning sense, we have to consider the labels provided by experts, usually considered as ground truth, to be noisy as well. Besides human error, also the difficulty of a certain data subset will play a role. The closer a single object is to a prototype, the easier the recognition for a human as well as for a machine learner, an effect which already successfully employed in other machine learning domains \cite{univis90537614}. 

Bengio \textit{et al.} suggest to exploit this difficulty in order to improve on generalization of machine learning approaches, effectively mimicking human learning, which is why they coined the term curriculum learning \cite{Bengio:2009dd}. While there are other approaches to provide this from the confidence of machine learning approaches inherently, a very strong indicator is human error. Multi label expert data sets provide thus the possibility to differentiate prototypical from difficult samples, while at the same time supplying the learning system with a better estimate for the ground truth. 

\subsection*{Existing software}

Besides commercially available digital histopathology software, e.g. provided by the manufacturers of the scanner hardware, there is a high number of open source software solutions available for slide viewing and analysis \cite{Bertram:2017kn}, with some products supporting whole-slide images and others using standard graphics formats. Many solutions provide not only annotation capabilities, but also plug-in systems for automated analysis or pre-processing (e.g. Icy \cite{deChaumont:2012iu} or CellProfiler \cite{Carpenter:2006iu}). As de Chaumont \textit{et al.} reported, the emphasis of these projects was collaborative design and evaluation of tools and algorithms within the bioinformatics community. 
To extend the collaborative approach to the annotation, Marée \textit{et al.} released Cytomine, an internet-based general-purpose annotation tool \cite{Maree:2013fp}, with full integration of whole-slide images and respective on-demand loading. Since within the tool, annotations are represented in layers, it is possible to do a blind annotation with multiple experts, which is valuable for shape estimations (e.g. for tumor regions) in a blinded manner. For cell type annotations, however, multiple opinions on one cell are of great importance to judge the prototypical character of an occurrence and hence the expected difficulty. 

In order to improve on current data sets, we can formulate requirements on software that would be used to aggregate data sets with multiple labels on the same cells and using big data amounts, which we do not find entirely fulfilled in state of the art solutions (see Table \ref{tab:comparison}):

\begin{itemize}
	\item The user interface must be intuitive to use and annotations can be set with very little interactions, using one click only if possible.
	\item The software must be able to support blinded class labeling, where the annotated cell is displayed but the label of previous annotators is hidden.
\end{itemize}

\begin{table*}[t]
\caption{Comparison of a selection of open source software packages suited for cell annotations.}
\centering
\resizebox{\textwidth}{!}{%
\begin{tabular}{lccccc}
\hline 
Product & Icy \cite{deChaumont:2012iu} & Qupath \cite{Bankhead:2017ex} &  Cytomine \cite{Maree:2013fp} & SlideRunner \\
\hline 
	
Whole-Slide Image support	&Yes (bio-formats)&	Yes (openslide)&	Yes (openslide)&	Yes (openslide)\\
Very large image support	&No (limit at 2.1 Gigapixels)	&Yes	&	Yes	&Yes\\
Multi-user Annotation	&No	&No	&Yes&	Yes\\
Single Click Cell Annotations	&No	&Yes&		No&	Yes\\
Blind Annotations	&No	&No&	Yes	&Yes\\
Blind Multi-labels	&No&	No	&	No	&Yes\\
Guided Screening Process & No & No & No & Yes \\
\hline 
\end{tabular}	
}
\label{tab:comparison}
\end{table*}

\section{Methods}
SlideRunner is a GPL-licensed tool\footnote{available at: \texttt{https://github.com/maubreville/SlideRunner}}, written in Python 3.5 and using OpenSlide \cite{Goode:2013dm} as image loading backend. It provides several modes of navigation, and two major modes of operation for annotation:
\begin{itemize}
	\item \texttt{Center annotation:} In this mode a single mouse click is needed to add an annotation in the center of an object, e.g. a cell.
	\item \texttt{Outline annotation:} Either using multiple clicks or click-and-drag, polygon curves can be added and annotated.
\end{itemize}

\begin{figure*}[b]
\centering
\subfigure[]{\label{gui}\includegraphics[width=0.5\textwidth]{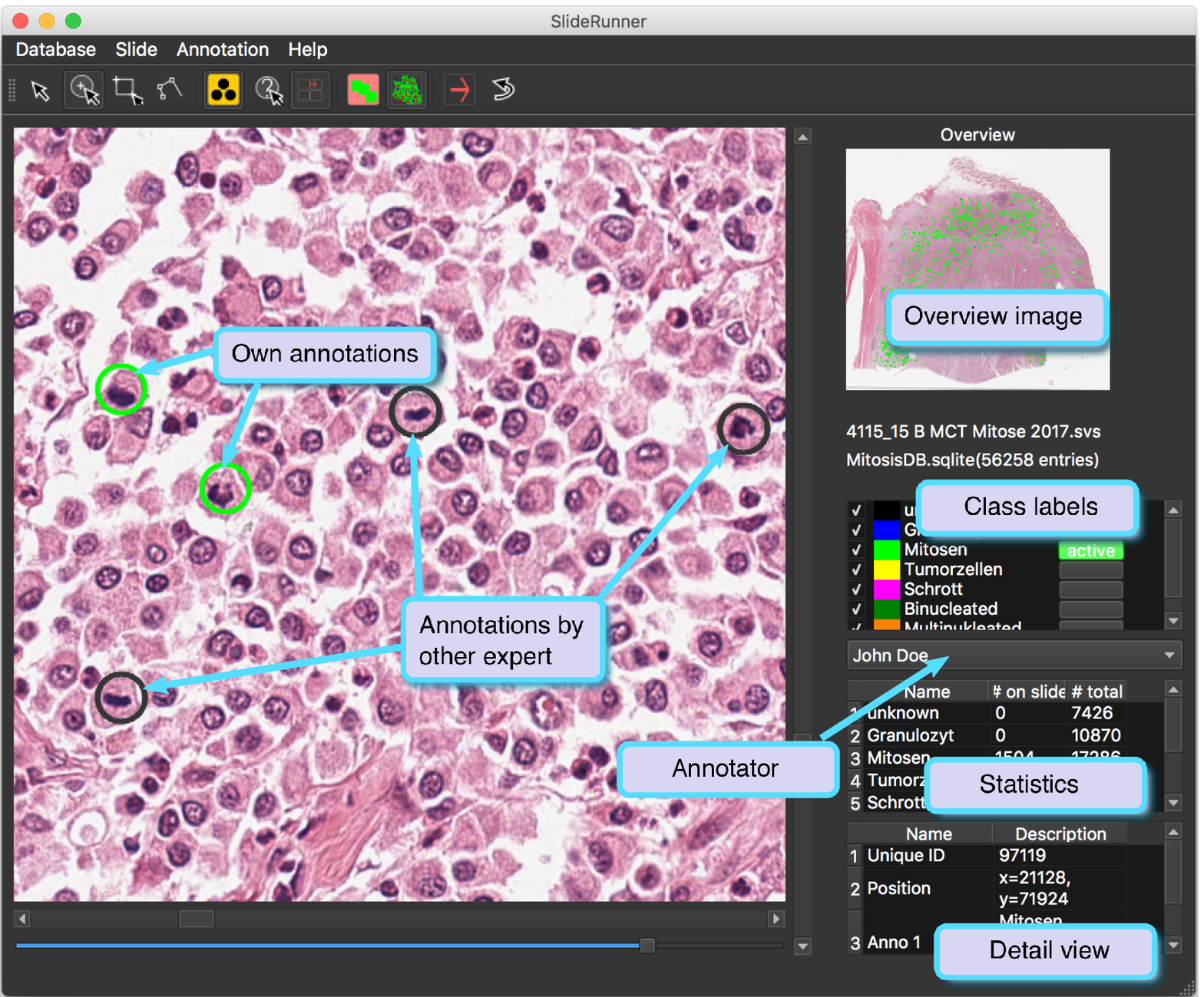}}
\subfigure[]{\label{uml}\includegraphics[width=0.45\textwidth]{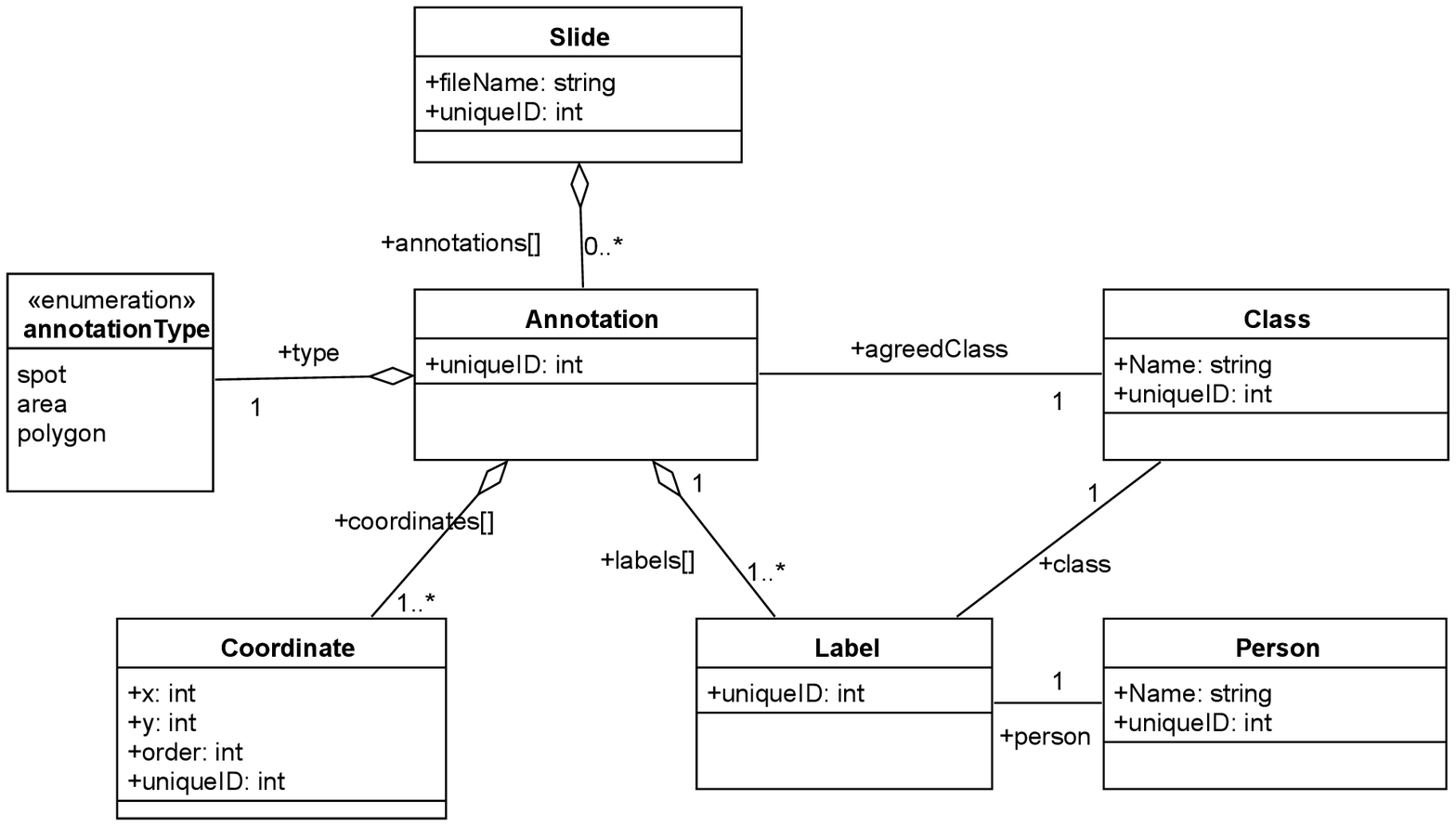}}
\caption{a) GUI overview in blinded mode. In this mode, annotations by other experts are only visible as unknown classes (in black). b) UML diagram of database structure. Every annotation may contain multiple labels from different persons and multiple coordinates.}

\end{figure*}

\subsection*{Data model}
As depicted in Fig. \ref{uml}, the major entity of our database model is the annotation. Each annotation can have multiple coordinates, with their respective x and y coordinates, and the order they were drawn (for polygons). Further, each annotation has a multitude of labels that were given by one person each and are belonging to one class, respectively. 

\subsection*{Blinded Annotation}

For blinded multi-annotation, it is important that the viewer is provided with a view that does present the area or object annotations of other experts, yet hides the information about the class the object was given but those other raters. For this, SlideRunner provides a mode where only own annotations are provided with (color-based) class informations in the image (see Fig. \ref{gui}). 

To leverage more gains in annotation performance, this mode is enriched by a \textit{discovery mode}, in which the user is automatically presented with a random new image section upon completion of the currently visible view until all annotated objects have been classified successfully. This mode is expected to considerably reduce required time to reach for the next unclassified cells to annotate.

\subsection*{Guided Screening}

\begin{figure*}[b]
\centering
\includegraphics[width=\textwidth]{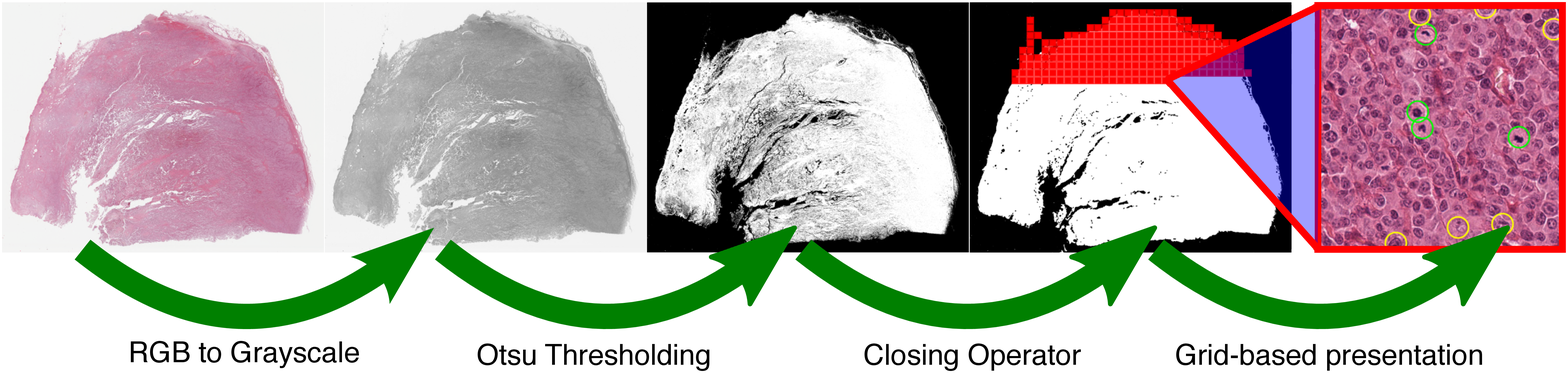}
\caption{Algorithmic toolchain for guided screening. Grid-based segments are presented from left to right, top to bottom.}
\label{screeningToolchain}
\end{figure*}

For the first annotation of the image, a guided screening mode is being provided that guides the expert at maximum optical zoom over the complete image. This mode will ensure that the observer will definitely examine every field of view of the image and is therefore especially helpful if complete annotation coverage is desired. In order to only display fields of interest containing tissue sections, white/empty areas need to be excluded from the image, which is performed by the algorithmic chain depicted in Fig. \ref{screeningToolchain}. After thresholding and morphologic closing, a grid is projected on the image, and non-empty slide partitions are shown to the user.

\section{Results}

With the means described in this text, a double-annotated database of sparsely annotated cell types from canine mast cell tumor slides was established. Histo-pathologically, canine cells show great similarity to human cells including the appearance and diversity of mitotic figures. As Table \ref{cm} shows, we found a good agreement between the raters in general, with Cohen's $\kappa = 0.815$.
Most disagreement can be found between clear decisions for one of the cell classes and the ambiguous class. For a curriculum learning-based approach, this may be a good hint towards the difficulty of detection. 
Mean annotation times (measured as time difference between annotation events) were $6.6\,s$ and $6.3\,s$ for both raters for first annotations, and $2.0\,s$ and $2.6\,s$ for second annotations, respectively (evaluated on N=71,561 labels).

\begin{table}[t]
\caption{Confusion matrix between two pathologists. Most significant disagreement was between the ambigous class and the clear classes.}
\centering
\begin{tabular*}{\textwidth}{l@{\extracolsep\fill}lllll}
\hline
 & granulocytes & mitotic cells & normal tumor cells & ambiguous \\
 \hline
granulocytes &  10318  &  395  &  327 &  2249 \\
mitotic cells &    147 &  30623 &  202 &   458 \\
normal tumor cells &     27 &   546 & 18445 &   387 \\
ambiguous &    257 &  2949 &  1331 &  2420 \\
\hline
\end{tabular*}	
\label{cm}
\end{table}

For one particular slide, one pathologist performed full annotation of mitotic figures manually, resulting in 2,252 single mitotic events distributed over the slide. For the same slide, a second pathologist performed an annotation using the guided screening mode, resulting in 4,233 mitotic events.

\section{Discussion} 
While minor differences were expected, the significant increase in mitotic figure annotations for usage of guided screening is surprising at first. Potentially, this effect can be attributed to a more thorough annotation in this mode, where the expert's attention is not focused on the center of the image. In general, statistics between both pathologists working in the same mode did not differ significantly for the whole data set.

We find that using the methods realized in SlideRunner lead to a fast annotation process of mitotic figures and other cell types. In part, however, this might also be related to a generally high mitotic count in the slides that have been labeled. In general, yet, using the tool provided means to build up a database of mitotic and other cell annotations that is unprecedented in size and could leverage precision gains provided by machine learning methods.

\bibliographystyle{model1-num-names}

\end{document}